\documentclass[10pt,twocolumn,letterpaper]{article}

\usepackage{cvpr}
\usepackage{times}
\usepackage{epsfig}
\usepackage{graphicx}
\usepackage{subcaption}
\usepackage{amsmath}
\usepackage{amssymb}
\makeatletter
\def\thickhline{%
  \noalign{\ifnum0=`}\fi\hrule \@height \thickarrayrulewidth \futurelet
   \reserved@a\@xthickhline}
\def\@xthickhline{\ifx\reserved@a\thickhline
               \vskip\doublerulesep
               \vskip-\thickarrayrulewidth
             \fi
      \ifnum0=`{\fi}}
\makeatother
\newlength{\thickarrayrulewidth}
\usepackage[pagebackref=true,breaklinks=true,letterpaper=true,colorlinks,bookmarks=false]{hyperref}

\cvprfinalcopy
\def\cvprPaperID{2635} % *** Enter the CVPR Paper ID here

\ifcvprfinal\fi
\begin{document}

%%%%%%%%% TITLE
\title{Recurrent Neural Network for (Un-)supervised Learning of Monocular Video Visual Odometry and Depth}

\author{Rui Wang, Stephen M. Pizer, Jan-Michael Frahm\\
University of North Carolina at Chapel Hill\\
}

\maketitle
%\thispagestyle{empty}

%%%%%%%%% ABSTRACT
\begin{abstract}

Deep learning-based, single-view depth estimation methods have recently shown highly promising results. However, such methods ignore one of the most important features for determining depth in the human vision system, which is motion. We propose a learning-based, multi-view dense depth map and odometry estimation method that uses Recurrent Neural Networks (RNN) and trains utilizing multi-view image reprojection and forward-backward flow-consistency losses. Our model can be trained in a supervised or even unsupervised mode. It is designed for depth and visual odometry estimation from video where the input frames are temporally correlated. However, it  also generalizes to single-view depth estimation. Our method produces superior results to the state-of-the-art approaches for single-view and multi-view learning-based depth estimation on the KITTI driving dataset. 
    
\end{abstract}

%%%%%%%%% BODY TEXT
\section{Introduction}
\label{sec:intro}

The tasks of depth and odometry (also called ego-motion) estimation are longstanding tasks in computer vision providing valuable information for a wide variety of tasks, \eg autonomous driving, AR/VR applications, and virtual tourism. %visual simultaneous localization and mapping (visual-SLAM), structure from motion (SfM) and multi-view stereo (MVS). 

\begin{figure}[ht!]
    \centering
    \includegraphics[width=\columnwidth]{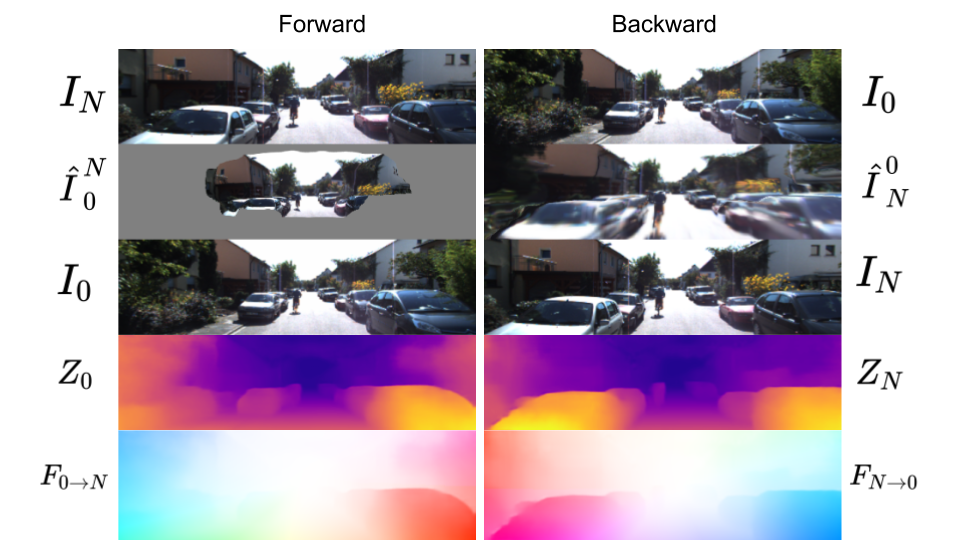}
    \caption{ Example results from our method. The first row shows the source image. The second row illustrates the projection of the source image into the target image. The third row shows the target image. The fourth row illustrates the estimated depth map and the last row illustrates the estimated optical flow. }
    \label{fig:teasor}
\end{figure}

Recently, convolutional neural networks (CNN) \cite{liu2015deep,eigen2015predicting,garg2016unsupervised,zhou2017unsupervised,ummenhofer2017demon} have begun to produce results of comparable quality to traditional geometric computer vision methods for depth estimation in measurable areas and achieve significantly more complete results for ambiguous areas through the learned priors. However, in contrast to traditional methods, most  CNN methods treat depth estimation as a single view task and thus ignore the important temporal information in monocular or stereo videos. The underlying rationale of these single view depth estimation methods is the possibility of human depth perception from a single image. However, they neglect the fact that motion is actually more important for the human to infer distance \cite{rogers1979motion}. We are constantly exposed to moving scenes, and the speed of things moving in the image is related to the combination of their relative speed and effect inversely proportional to their depth.

%By combining single image depth estimation with two-view or multi-view visual odometry, \cite{zhou2017unsupervised,Yin_2018_CVPR,Zhan_2018_CVPR} proposed unsupervised depth and odometry estimation pipelines driven by image reprojection error.

In this work, we propose a framework that simultaneously estimates the visual odometry and depth maps from a
video  sequence  taken  by  a  monocular  camera. To be more specific, we use convolutional Long Short-Term Memory (ConvLSTM) \cite{xingjian2015convolutional} units to carry temporal information from previous views into the current frame's depth and visual odometry estimation. We have improved upon existing deep single- and two-view stereo depth estimation methods by interleaving  ConvLSTM units with the convolutional layers to effectively utilize multiple previous frames in each estimated depth maps. Since we utilize multiple views, the image reprojection constraint between multiple views can be incorporated into the loss, which shows significant improvements for both supervised and unsupervised depth and camera pose estimation. 

In addition to the image reprojection constraint, we further utilize a forward-backward flow-consistency constraint \cite{Yin_2018_CVPR}. Such a constraint provides additional supervision to image areas where the image reprojection is ambiguous. Moreover, it improves the robustness and generalizability of the model. Together these two constraints can even allow satisfactory models to be produced when groundtruth is unavailable at training time. Figure \ref{fig:teasor} shows an example of  forward-backward image reprojection and optical flow as well as the resulting predicted depth maps.

\begin{figure*}[h]
    \centering
\includegraphics[width=\textwidth,height=7cm]{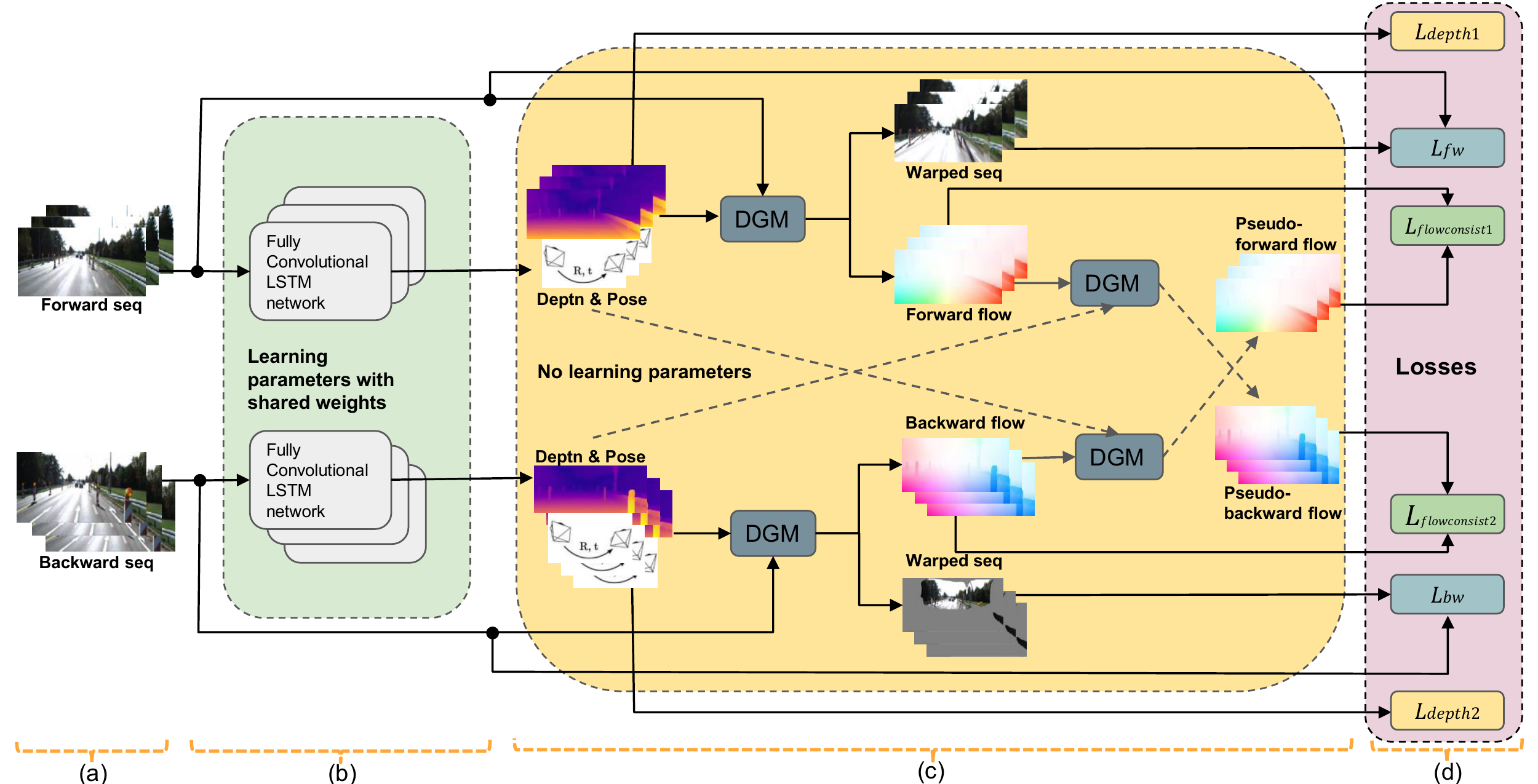}
       \caption{Training pipeline of our proposed RNN-based depth and visual odometry estimation network. During training our framework takes forward and backward 10-frame subsequences as input and uses multi-view image reprojection, flow-consistency, and optionally groundtruth depth to train our depth and visual odometry networks. DGM is a differentiable geometric module.}
    \label{pipeline}
\end{figure*}

We summarize our innovations as follows: 1) An RNN architecture for monocular depth and odometry estimation that uses multiple consecutive views. It does so by incorporating LSTM units, as used in natural language processing, into depth and visual odometry estimation networks. 2) These LSTM units importantly allow the innovation of using depth and camera motion estimation to benefit from the richer constraints of a multi-view process. In particular, they use multi-view image reprojection and forward-backward flow-consistency constraints to produce a more accurate and consistent model. 3) This design allows two novel capabilities: a) it can be trained in both supervised and unsupervised fashion; b) it can continuously run on arbitrary length sequences delivering a consistent scene scale.  

We demonstrate on the  KITTI \cite{Geiger2012CVPR} benchmark dataset that our method can produce superior results over the state-of-the-art for both supervised and unsupervised training. We will release source code upon acceptance. 
    
%\end{itemize}

%-------------------------------------------------------------------------
\section{Related work}

Traditionally, the 3D reconstruction and localization are mostly solved by pure geometric reasoning. SfM and SLAM are the two most prevalent frameworks for sparse 3D reconstruction of rigid geometry from images. SfM is typically used for offline 3D reconstruction from unordered image collections, while visual SLAM aims for a real-time solution using a single camera \cite{davison2007monoslam,newcombe2011dtam}. More recent works on SLAM systems include ORB-SLAM \cite{mur2015orb} and DSO \cite{engel2018direct}. Sch\"{o}nberger and Frahm \cite{schoenberger2016sfm} review the state-of-the-art in SfM and propose an improved incremental SfM method.  

Recently, CNNs are increasingly applied to 3D reconstruction, in particular, to the problem of 3D reconstruction of dense monocular depth,
%. Dense monocular depth estimation has gained interest because regressing the depth representation 
which is similar to the segmentation problem and thus the structure of the CNNs can be easily adapted to the task of depth estimation \cite{long2015fully}.

\textbf{Supervised methods.} Eigen \textit{et al.} \cite{eigen2015predicting} and Liu \textit{et al.} \cite{liu2015deep} proposed end-to-end networks for single-view depth estimation, which opened the gate for deep learning-based supervised single-view depth estimation. Following their work, Laina \textit{et al}. \cite{laina2016deeper} proposed a deeper residual network for the same task. Qi \textit{et al.} \cite{Qi_2018_CVPR} jointly predicted depth and surface normal maps from a single image. Fu \textit{et al.} \cite{Fu_2018_CVPR} further improved the network accuracy and convergence rate by learning it as an ordinal regression problem. Li \textit{et al.} \cite{Li_2018_CVPR} used modern structure-from-motion and multi-view stereo (MVS) methods together with multi-view Internet photo collections to create the large-scale %dataset called 
MegaDepth dataset providing improved depth estimation accuracy via bigger training dataset size. We improve upon these single-view methods by utilizing multiple views through an RNN architecture %, which we show 
to generate more accurate depth and pose. %estimation results.

Two-view or multi-view stereo methods have traditionally been the most common techniques for dense depth estimation. For the interested reader, Scharstein and Szeliski \cite{scharstein2002taxonomy} give a comprehensive review on two-view stereo methods. Recently, Ummenhofer \textit{et al}. \cite{ummenhofer2017demon} formulated two-view stereo as a learning problem. They showed that by explicitly incorporating dense correspondences estimated from optical flow into the two-view depth estimation, they can force the network to utilize stereo information on top of the single view priors. There is currently a very limited body of CNN-based multi-view reconstruction methods. Choy \textit{et al}. \cite{choy20163d} use an RNN to reconstruct the object in the form of a 3D occupancy grid from multiple viewpoints. Yao \textit{et al.} \cite{Yao_2018_ECCV} proposed an end-to-end deep learning framework for depth estimation from multiple views. They use differentiable homography warping to build a 3D cost volume from one reference image and several source images. Kumar \textit{et al.} \cite{kumar2018depthnet} proposed an RNN architecture that can learn depth prediction from monocular videos. However, their simple training pipeline, e.g., no explicit temporal constraints, failed to explore the full capability of the network. Our method is trained with more sophisticated multi-view reprojection losses and can perform both single-view and multi-view depth estimation. 

\begin{figure}[ht]
    \centering
    \includegraphics[width=\columnwidth]{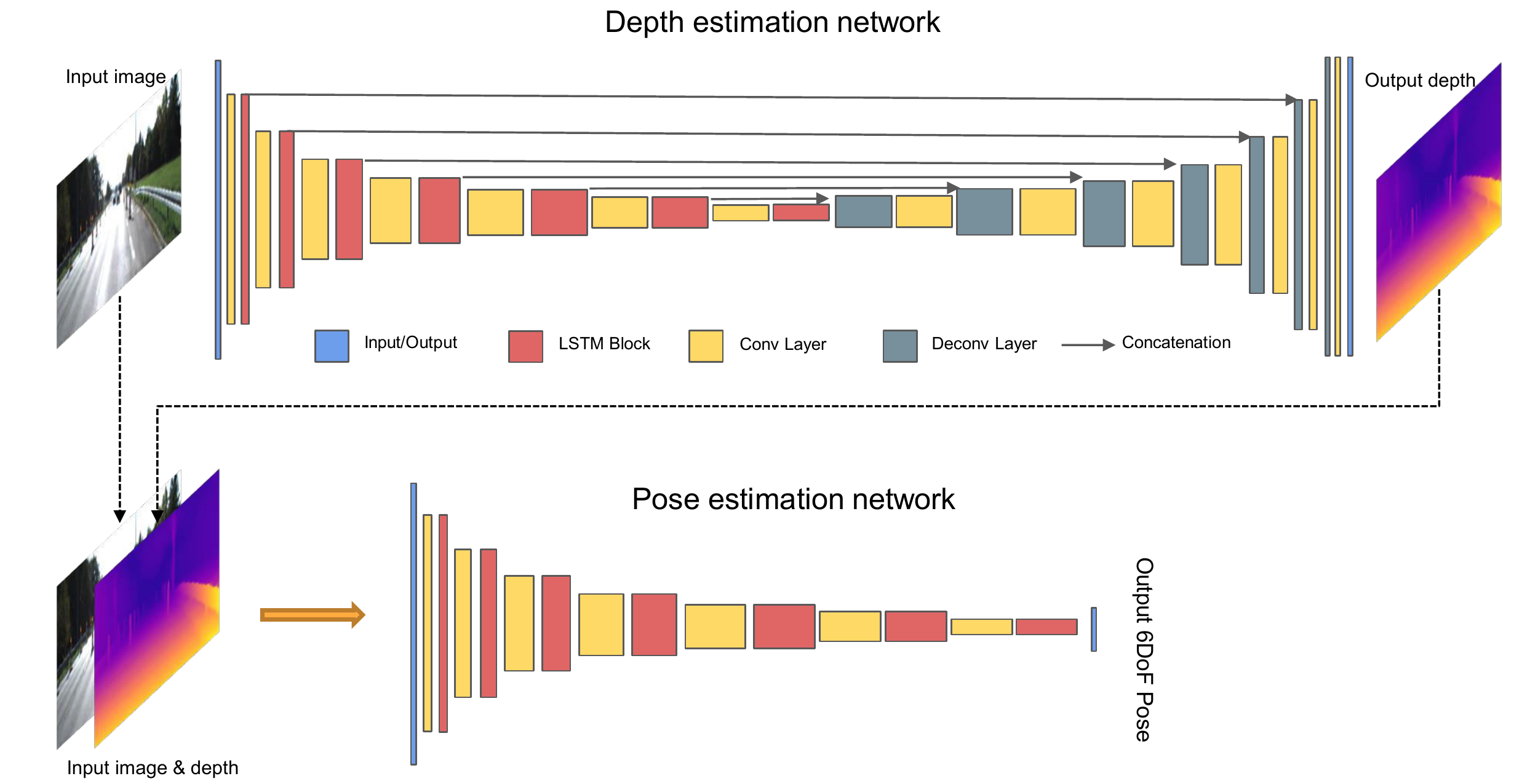}
    \caption{Overall network architecture of our RNN-based depth and visual odometry estimation framework.  The height of each rectangle represents the size of its feature maps, where each smaller feature map is half the size of the preceding feature map.}
    \label{fig:network}
\end{figure}

\textbf{Unsupervised methods.} Recently, by incorporating elements of view synthesis \cite{zhou2016view} and Spatial Transformer Networks \cite{jaderberg2015spatial}, monocular depth estimation has been trained in an unsupervised fashion. This was done by transforming the depth estimation problem into an image reconstruction problem where the depth is the intermediate product that integrates into the image reconstruction loss. Godard \textit{et al.}\cite{godard2017unsupervised}, and Garg \textit{et al.}\cite{garg2016unsupervised} use stereo pairs to train CNNs to estimate disparity maps from single views. Luo \textit{et al.} \cite{Luo_2018_CVPR} leverage both stereo and temporal constraints to generate improved depth at known scale. Zhou \textit{et al.}
\cite{zhou2017unsupervised} further relax the needs of stereo images to monocular video by combining a single view depth estimation network with a multi-view odometry estimation network. Following Zhou \textit{et al.} \cite{zhou2017unsupervised}'s work, Mahjourian \textit{et al.} \cite{Mahjourian_2018_CVPR} further enforced consistency of the estimated 3D point clouds and ego-motion across consecutive frames. In addition to depth and ego-motion, Yin \textit{et al.} \cite{Yin_2018_CVPR} also jointly learn optical flow in an end-to-end manner which imposed additional geometric constraints. However, due to scale ambiguity and the lack of temporal constraints, these methods cannot be directly applied for full trajectory estimation on monocular videos. By leveraging recurrent units, our method can run on arbitrary length sequences delivering a consistent scene scale.

%-------------------------------------------------------------------------

\section{Method}

In this section we introduce our method for multi-view depth and visual odometry estimation. We first describe our recurrent neural network architecture and then the multi-view reprojection and forward-backward flow-consistency constraints for the network training.  

\subsection{Network Architecture}

Our architecture, shown in Figure \ref{fig:network}, is made up of two networks, one for depth and one for visual odometry. 
%We incorporate recurrent units into a CNN to leverage temporal information from one or multiple previous views into current view depth and camera pose estimation, making it more accurate for continuous video sequences. 

\textbf{Our depth estimation network} uses a U-shaped network architecture similar to DispNet \cite{mayer2016large}. Our main innovation is to interleave recurrent units into the encoder which allows the network to leverage not only spatial but also temporal information in the depth estimation. The spatial-temporal features computed by the encoder are then fed into the decoder for accurate depth map reconstruction. The ablation study in Secion \ref{sec:abl} confirms our choice for the placements of the ConvLSTM \cite{xingjian2015convolutional} units. %The lines connecting corresponding layers in the encoder and decoder are skip-connections.
Table \ref{table:netparam} shows the detailed network architecture. The input to the depth estimation network is a single RGB frame $I_t$ and the hidden states $h_{t-1}^d$ from the previous time-step ($h_{t-1}^d$ are initialized to be all zero for the first time-step). The hidden states are transmitted internally through the ConvLSTM units. The output of our depth estimation network are the depth map $Z_t$ and the hidden states $h_t^d$ for the current time-step . %The down-sampling from a previous layer to the next is done by a stride-2 convolution instead of max-pooling. 

\begin{table}[h!]
\centering
\small
\resizebox{\linewidth}{!}{
\begin{tabular}{ p{3cm}| p{2.5cm}|p{2.5cm} }
 \hline
 Type & Filters & Output size \\
 \hline
 Input &  &  128 $\times$416$\times$3\\
Conv+ConvLSTM &32@3$\times$3$\times$3 & 64$\times$208$\times$32 \\
Conv+ConvLSTM &64@3$\times$3$\times$32 & 32$\times$104$\times$64\\ 
Conv+ConvLSTM &128@3$\times$3$\times$64  & 16$\times$52$\times$128\\
Conv+ConvLSTM &256@3$\times$3$\times$128 & 8$\times$26$\times$256\\
Conv+ConvLSTM &256@3$\times$3$\times$256  & 4$\times$13$\times$256\\
Conv+ConvLSTM &256@3$\times$3$\times$256  & 2$\times$7$\times$256\\
Conv+ConvLSTM &512@3$\times$3$\times$256  & 1$\times$4$\times$512\\
\hline
Deconv+Concat+Conv &256@3$\times$3$\times$512 & 2$\times$7$\times$256\\ 
Deconv+Concat+Conv &128@3$\times$3$\times$256  & 4$\times$13$\times$128\\
Deconv+Concat+Conv &128@3$\times$3$\times$128 & 8$\times$26$\times$128\\
Deconv+Concat+Conv &128@3$\times$3$\times$128  & 16$\times$52$\times$128\\
Deconv+Concat+Conv &64@3$\times$3$\times$128  & 32$\times$104$\times$64\\
Deconv+Concat+Conv &32@3$\times$3$\times$64  & 64$\times$208$\times$32\\
Deconv &16@3$\times$3$\times$32  & 128$\times$416$\times$16\\
Conv (output) &1@3$\times$3$\times$16  & 128$\times$416$\times$1\\
\hline
\end{tabular}
}
\caption{Detailed depth estimation network architecture. Every convolution in the encoder uses stride 2 for downsampling. Before the output a sigmoid activation function is used to ensure the output is in range [0, 1]; All the other convolutions and decovolutions are followed by batch norm and LeakyRELU activation. }
\label{table:netparam}
\end{table}

%For the encoder each step is a composition of two convolutions. The first one is a 3$\times$3 convolution with stride 2 for down-sampling, and the second one is a 3$\times$3 dilated convolution with dilation 2 and stride 1. For the decoder each step is a combination of a deconvolution, a concatenation and a convolutional LSTM. The deconvolution is 3$$\times$$3 with stride 2. The concatenation combines the upsampled feature map with the corresponding feature map in the encoder. The convolutional LSTM takes the concatenated feature map together with the hidden state from the previous time-step and outputs a new feature map as well as a new hidden state. 

%Before the output layer a sigmoid activation function is used to ensure the output is in range [0, 1];  all the other convolutions and deconvolutions are followed by leaky rectified linear units (LeakyRELU). Since we are estimating the inverse depth, a small constant $\varepsilon$ is added to the output so that the network will not output depth at infinity.

%Our depth estimation network, even the unsupervised version, can run on an arbitrary length video sequence with the prediction at a single scale.

\textbf{ Our visual odometry network} uses a VGG16 \cite{simonyan2014very} architecture with recurrent units interleaved. Table \ref{table:odomnet} shows the detailed network architecture. The input to our visual odometry network is the concatenation of $I_t$ and $Z_t$ together with the hidden states $h_{t-1}^p$ from the previous time-step. The output is the relative 6DoF camera pose $P_{t\rightarrow t-1}$ between the current view and the immediately preceeding view. The main differences between our visual odometry network and most current deep learning-based visual odometry methods are 1) At each time-step, instead of a stack of frames, our visual odometry network only takes the current image as input; the knowledge about previous frames is in the hidden layers.     
2) Our visual odometry network also takes the current depth estimation as input, which ensures a consistent scene scale between depth and camera pose (important for unsupervised depth estimation, where the scale is ambiguous). 3) Our visual odometry network can run on a full video sequence while maintaining a single scene scale. % Figure \ref{fig:traj_scale} shows that our method can estimate a full trajectory at a single scale.  

\begin{table}[t!]
\centering
\small
\resizebox{\linewidth}{!}{
\begin{tabular}{ p{3cm}| p{2.5cm}|p{2.5cm} }
 \hline
 Type & Filters & Output size \\
 \hline
 Input &  &  128$\times$416$\times$4\\
Conv+ConvLSTM&32@3$\times$3$\times$3 & 64$\times$208$\times$32 \\
Conv+ConvLSTM &64@3$\times$3$\times$32 & 32$\times$104$\times$64\\ 
Conv+ConvLSTM &128@3$\times$3$\times$64  & 16$\times$52$\times$128\\
Conv+ConvLSTM &256@3$\times$3$\times$128 & 8$\times$26$\times$256\\
Conv+ConvLSTM &256@3$\times$3$\times$256  & 4$\times$13$\times$256\\
Conv+ConvLSTM &256@3$\times$3$\times$256  & 2$\times$7$\times$256\\
Conv+ConvLSTM &512@3$\times$3$\times$256  & 1$\times$4$\times$512\\
Conv (output) &6@1$\times$1$\times$ 512  & 1$\times$ 1$\times$6\\
\hline
\end{tabular}
}
\caption{Detailed visual odometry network architecture.  Every convolution (except for output layer) is followed by batch normalization and RELU as activation.  }
\label{table:odomnet}
\end{table}
%Each step contains 3$\times$3 convolution with stride 2 for downsampling and a convolutional LSTM to take information from previous time-steps and output hidden state to carry into the next time-step. Every convolution except at the output layer is followed by a rectified linear unit (RELU) as activation function. 

\subsection{Loss Functions}

\subsubsection{Multi-view Reprojection Loss}
\label{mtv loss}
Zhou \textit{et al.} \cite{zhou2017unsupervised} showed that the learning of depth and visual odometry estimation can be formulated as an image reconstruction problem using a differentiable geometric module (DGM). Thus we can use the DGM to formulate an image reconstruction constraint between $I_t$ and $I_{t-1}$ using the estimated depth $Z_t$ and camera pose $P_{t \rightarrow t-1}$ as introduced in the previous subsection. However, such a pairwise photometric consistency constraint is very noisy due to illumination variation, low texture, occlusion, etc. Recently, Iyer \textit{et al.} \cite{iyer2018geometric} proposed a composite transformation constraint for self-supervised visual odometry learning. By combining the pairwise image reconstruction constraint with the composite transformation constraint, we propose a multi-view image reprojection constraint that is robust to noise and provides strong self-supervision for our multi-view depth and visual odometry learning. As shown in Figure \ref{pipeline}(c), the output depth maps and relative camera poses together with the input sequence are fed into a differentiable geometric module (DGM) that performs differentiable image warping of every previous view of the sub-sequence into the current view. Denote the input image sequence (shown in Figure \ref{pipeline}(a)) as $\{I_t|t=0...N-1\}$, the estimated depth maps as 
$\{Z_t|t=0...N-1\}$, and the camera poses as the transformation matrices from frame $t$ to %frame 
$t-1$: $\{P_{t \rightarrow t-1}|t=0...N-1$\}. The multi-view reprojection loss is
\begin{equation} 
L_{fw} = \sum_{t=0}^{N-1}\sum_{i=0}^{t-1} \sum_{\Omega} \lambda_t^i \, \omega_t^i \,|I_t-\hat{I}_t^i|
\label{eq:mtv_proj}
\end{equation}
where $\hat{I}_t^i$ is the $i^{th}$ view warped into $t^{th}$ view, $\Omega$ is the image domain, $\omega_t^i$ is a binary mask indicating whether a pixel of $I_t$ has a counterpart in $I_i$, and $\lambda_t^i$ is a weighting term that decays exponentially based on $t-i$. Image pairs that are far away naturally suffer from larger reprojection error due to interpolation and moving foreground so we use $\lambda_t^i$ to reduce the effect of such artifacts.  $\omega_t^i$ and $\hat{I}_t^i$ are obtained as 

\begin{equation} 
\omega_t^i, \hat{I}_t^i, F_{t\rightarrow i} = \phi(I_{i}, Z_t, P_{t\rightarrow i}, K)
\label{eq:phi}
\end{equation}
where $F_{t\rightarrow i}$ is a dense flow field for 2D pixels from view $t$ to view $i$, which is used to compute flow consistency. $K$ is the camera intrinsic matrix. The pose change from view $t$ to $i$, $P_{t\rightarrow i}$ can be obtained by a composite transformation as 
\begin{equation} 
P_{t\rightarrow i} = P_{i+1\rightarrow i} \cdot ... \cdot P_{t-1\rightarrow t-2} \cdot P_{t\rightarrow t-1}
\end{equation}

The function $\phi$ in Equation \ref{eq:phi} warps image $I_{i}$ into $I_{t}$ using $Z_t$ and $P_{t\rightarrow i}$. The function $\phi$ is a DGM \cite{Zhan_2018_CVPR}, which performs a series of differentiable 2D-to-3D, 3D-to-2D projections, and bi-linear interpolation operations \cite{jaderberg2015spatial}.  

In the same way, we reverse the input image sequence and perform another pass of depth $\{Z_t|t=N-1...0\}$ and camera pose $\{P_{t\rightarrow t+1}|t=N-1...0\}$ estimation,  obtaining the backward multi-view reprojection loss $L_{bw}$. This multi-view reprojection loss can fully exploit the temporal information in our ConvLSTM units from multiple previous views by explicitly putting constraints between the current view and every previous view. %Through these constraints the networks are forced to explore the temporal information in the ConvLSTM units. 

A trivial solution to Equation \ref{eq:mtv_proj} is  $\omega_t^i$ to be all zeros. To prevent the network from converging to the trivial solution, we add a regularization loss $L_{reg}$ to $\omega_t^i$, which gives a constant penalty to locations where $\omega_t^i$ is zero.

\subsubsection{Forward-backward Flow Consistency Loss}

A forward-backward consistency check has become a popular strategy in many learning-based tasks, such as optical flow\cite{hur2017mirrorflow}, registration
\cite{zhang2018inverse}, and depth estimation \cite{Yin_2018_CVPR,godard2017unsupervised,vijayanarasimhan2017sfm}, which provides additional self-supervision and regularization. Similar to \cite{Yin_2018_CVPR,vijayanarasimhan2017sfm}
 we use the dense flow field as a hybrid forward-backward consistency constraint for both the estimated depth and pose. We first introduce a forward-backward consistency constraint on a single pair of frames and then generalize to a sequence. Let us denote a pair of consecutive frames as $I_A$ and $I_B$, and their estimated depth maps and relative poses as $Z_A$, $Z_B$, $P_{A\rightarrow B}$, and $P_{B\rightarrow A}$. We can obtain a dense flow field $F_{A \rightarrow B}$ from frame $I_A$ to $I_B$ using Equation \ref{eq:phi}. Similarly we can obtain $F_{B \rightarrow A}$ using $Z_B$, $P_{B\rightarrow A}$. Using $F_{B \rightarrow A}$ we can compute a pseudo-inverse flow $\hat{F}_{A\rightarrow B}$ (due to occlusion and interpolation) as
 %where $P_{A\rightarrow B}$ is the relative pose from frame $I_A$ to $I_B$.
\begin{equation} 
\omega_A^B, \hat{F}_{A\rightarrow B}, F_{A\rightarrow B} = \phi(-F_{B\rightarrow A}, Z_A, P_{A\rightarrow B}, K)
\label{eq:pseudo_inverse}
\end{equation}
This is similar to Equation \ref{eq:phi} except that we are interpolating $F_{A\rightarrow B}$ from $-F_{B\rightarrow A}$ instead of $I_t$ from $I_i$. Therefore, we can formulate the flow consistency loss as 
\begin{equation} 
L_{flowconsit} = \omega_A^B \cdot |F_{A\rightarrow B}-\hat{F}_{A\rightarrow B}|+\omega_B^A \cdot |F_{B\rightarrow A}-\hat{F}_{B\rightarrow A}|
\label{eq:consistency}
\end{equation}
This is performed for every consecutive pair of frames in the input sequence. Unlike the multi-view reprojection loss we only compute flow-consistency on pairs of consecutive frames given the fact that the magnitude of the flow increases, for frame pairs that are far apart, leading to inaccurate pseudo-inverses due to interpolation.  

\subsubsection{Smoothness Loss}

Local smoothness is a common assumption for depth estimation. Following Zhan \textit{et al.} \cite{Zhan_2018_CVPR}, we use an edge-aware smoothness constraint which is defined as
\begin{equation} 
L_{smooth} = \sum_{t=0}^{N-1}\sum_{\Omega}|\nabla \xi_t| \cdot e^{-|\nabla I_t|}
\label{eq:smooth}
\end{equation}
where $\xi_t$ is the inverse depth. 

\subsubsection{Absolute depth loss}

The combination of multi-view reprojection loss $L_{fw}$, $L_{bw}$ defined in Equation \ref{eq:mtv_proj}, forward-backward flow-consistency loss $L_{flowconsist}$ defined in Equation \ref{eq:consistency}, and smoothness loss $L_{smooth}$ defined in Equation \ref{eq:smooth} can form an unsupervised training strategy for the network. This manner of training is suitable for cases where there is no groundtruth depth available, which is true for the majority of real world scenarios. However, the network trained in this way only produces depth at a relative scale. So optionally, if there is groundtruth depth available, even sparsely, we can train a network to estimate depth at absolute scale by adding the absolute depth loss defined as
\begin{equation} 
L_{depth} = \sum_{t=0}^{N-1}\sum_{\Omega}|\xi_t-\hat{\xi_t}|
\end{equation}
In addition, we can replace the local smoothness loss in Equation \ref{eq:smooth} by a gradient similarity to the groundtruth depth, which can be defined as
\begin{equation} 
L_{smooth} = \sum_{t=0}^{N-1}\sum_{\Omega}|\nabla \xi_t-\nabla\hat{\xi_t}|
\end{equation}

\subsection{Training Pipeline}

The full training pipeline of our method is shown in Figure \ref{pipeline}. Every $N$ consecutive key frames (we use $N=10$ in all our experiments)  are grouped together as an input sequence $S_{fw}$. The frames are grouped in a sliding window fashion such that more training data can be generated. Here the key frame selection is based on the motion between successive frames. Because the image reprojection constraints are ambiguous for very small baselines, we discard frames with baseline motion smaller than $\sigma$. Before passing the sequence to the network for training, we also reverse the sequence to create a backward sequence $S_{bw}$, which not only serves as a data augmentation but also is used to enforce the forward-backward constraints. The input sequence $S_{fw}$ is generated offline during the data preparation stage while the backward sequence $S_{bw}$ is generated online during the data preprocessing stage. $S_{fw}$ and $S_{bw}$ are fed into  two networks with shared weights; each generates a sequence of depth maps and camera poses as shown in Figure \ref{pipeline}. The estimated depth maps and camera poses are then utilized to generate dense flows to warp previous views to the current view through a differentiable geometric module (DGM) \cite{zhou2016view,Yin_2018_CVPR}. Furthermore, we utilize DGMs to generate the pseudo-inverse flows for both the forward and backward flows. By combining image warping loss, flow-consistency loss, and optionally absolute depth loss, we form the full training pipeline for our proposed framework. 

Once trained, our framework can run on arbitrary length sequences without grouping frames into fixed length sub-sequences. To bootstrap the depth and pose estimation, the hidden states for the ConvLSTM units are initialized by zero for the first frame. All following estimations will then depend on the hidden states from the previous  time-step.

\begin{figure*}[ht!]
\captionsetup[subfigure]{labelformat=empty}
  \centering
  \begin{subfigure}[b]{0.12\linewidth}
  \caption{\scriptsize Input}
  \includegraphics[width=\linewidth]{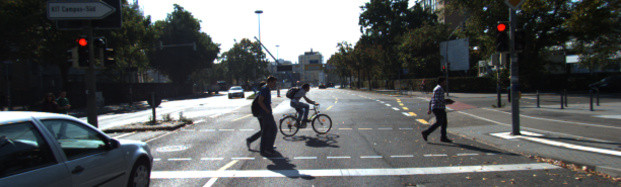}
  \end{subfigure}
  \begin{subfigure}[b]{0.12\linewidth}
  \caption{\scriptsize GT}
    \includegraphics[width=\linewidth]{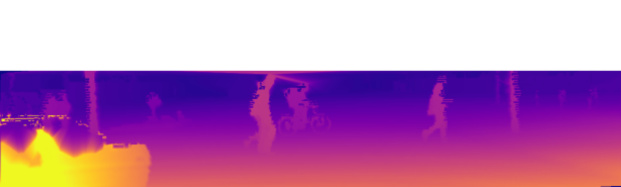}
  \end{subfigure}
    \begin{subfigure}[b]{0.12\linewidth}
  \caption{\scriptsize Yang \textit{et al}\cite{yang2018deep}}
    \includegraphics[width=\linewidth]{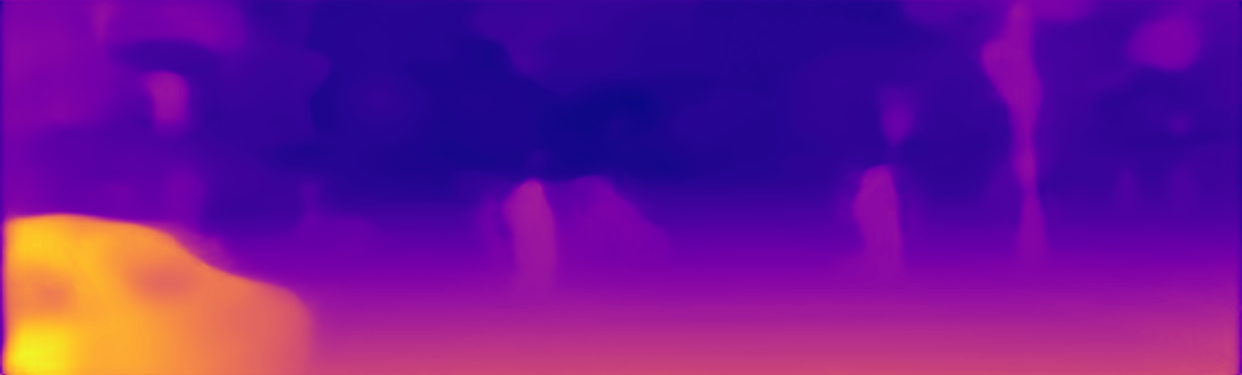}
  \end{subfigure}
    \begin{subfigure}[b]{0.12\linewidth}
  \caption{\scriptsize Kuznietsov \textit{et al.}\cite{kuznietsov2017semi}}
    \includegraphics[width=\linewidth]{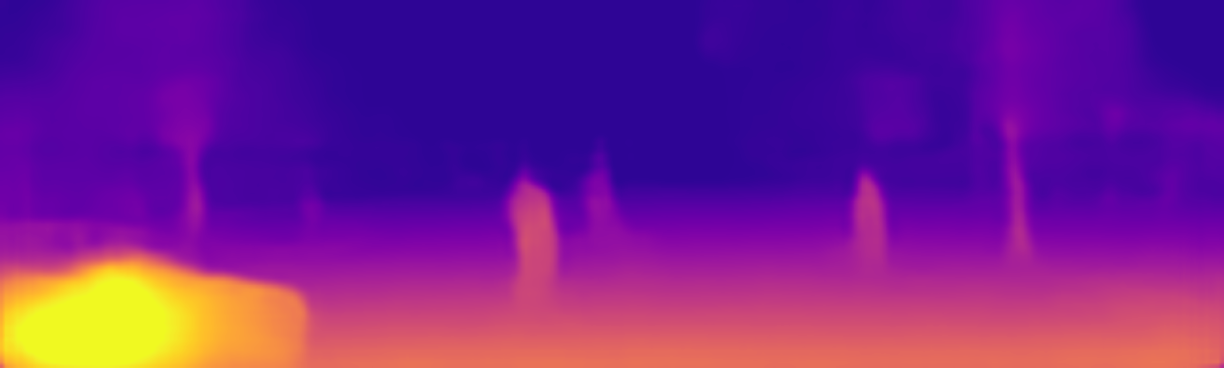}
  \end{subfigure}
    \begin{subfigure}[b]{0.12\linewidth}
  \caption{\scriptsize Godard \textit{et al.} \cite{godard2017unsupervised}}
  \includegraphics[width=\linewidth]{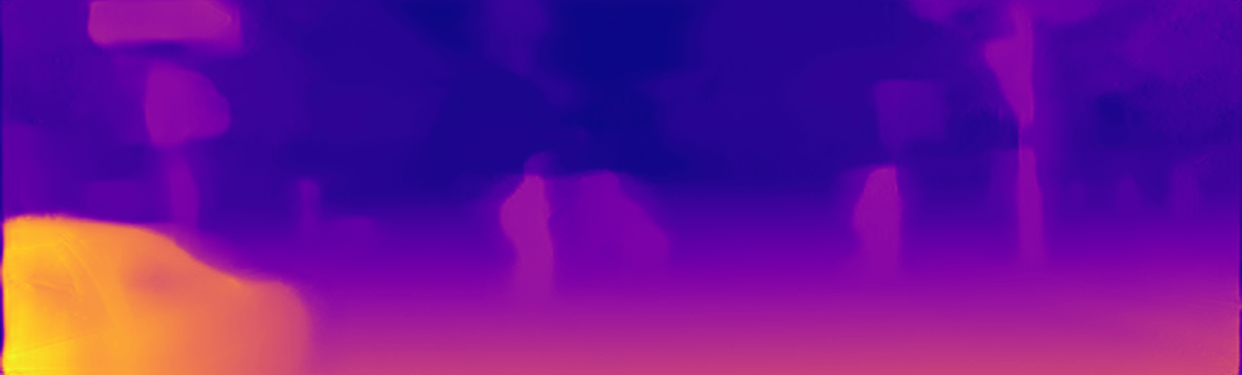}
  \end{subfigure}
    \begin{subfigure}[b]{0.12\linewidth}
  \caption{\scriptsize Garg \textit{et al.} \cite{garg2016unsupervised}}
  \includegraphics[width=\linewidth]{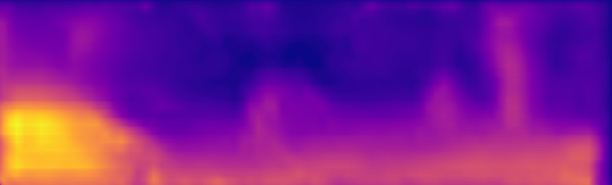}
  \end{subfigure}
    \begin{subfigure}[b]{0.12\linewidth}
  \caption{\scriptsize Ours unsup}
  \includegraphics[width=\linewidth]{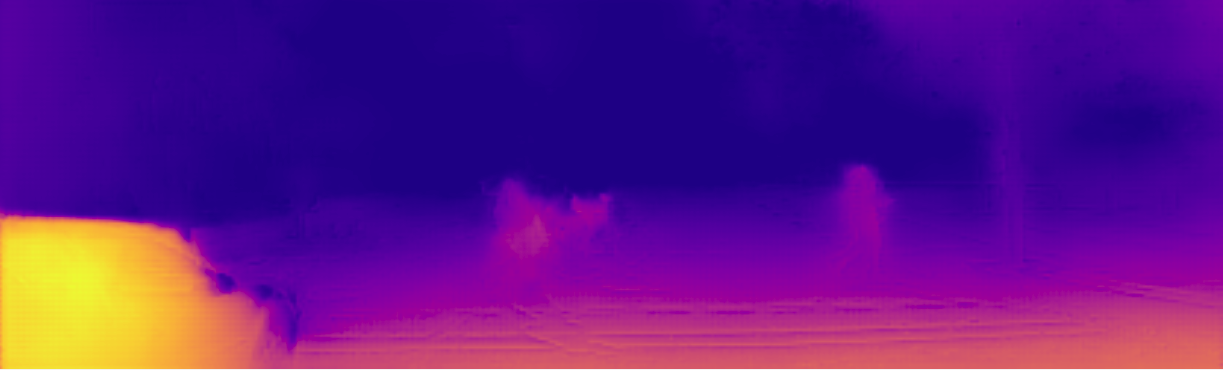}
  \end{subfigure}
    \begin{subfigure}[b]{0.12\linewidth}
  \caption{\scriptsize Ours sup}
  \includegraphics[width=\linewidth]{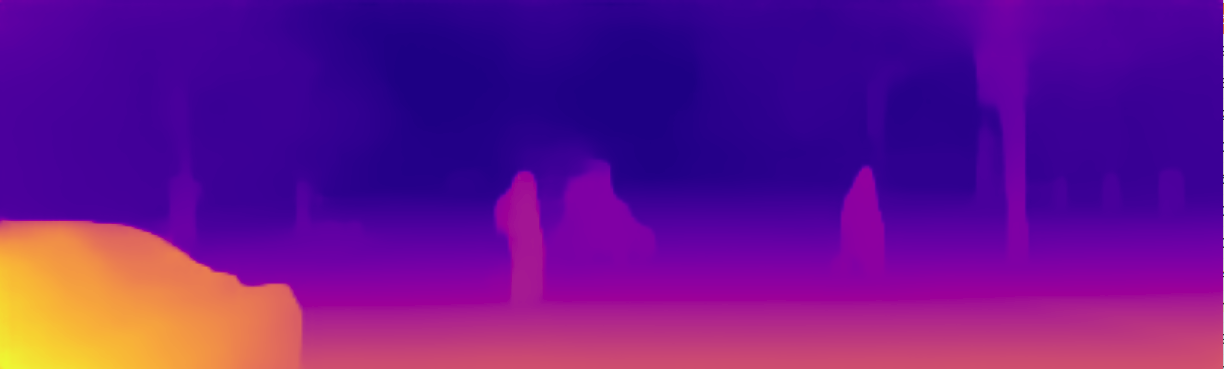}
  \end{subfigure}

  \begin{subfigure}[b]{0.12\linewidth}
  \includegraphics[width=\linewidth]{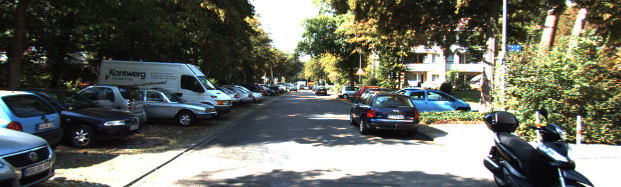}
  \end{subfigure}
  \begin{subfigure}[b]{0.12\linewidth}
    \includegraphics[width=\linewidth]{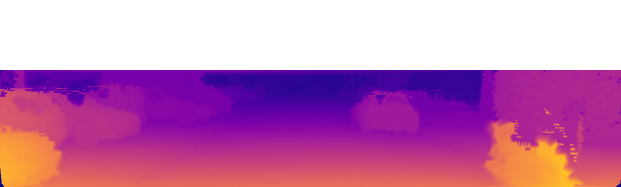}
  \end{subfigure}
    \begin{subfigure}[b]{0.12\linewidth}
    \includegraphics[width=\linewidth]{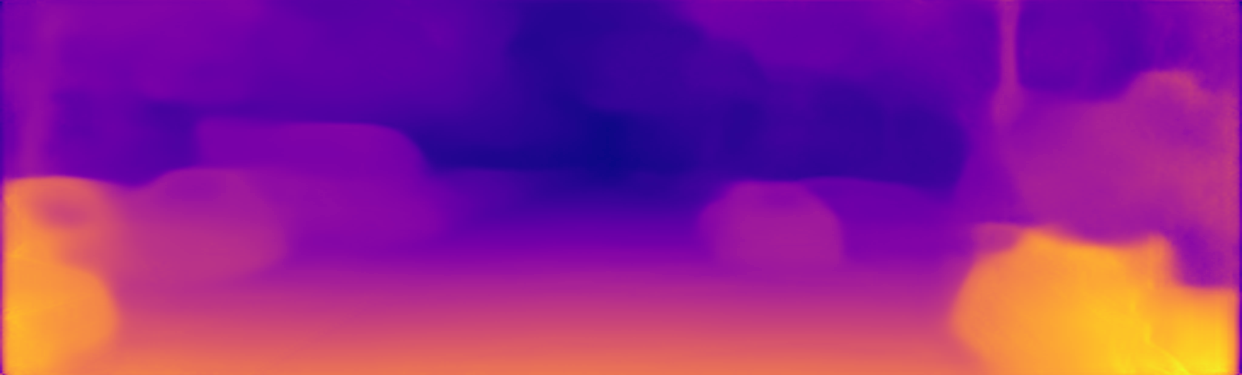}
  \end{subfigure}
    \begin{subfigure}[b]{0.12\linewidth}
    \includegraphics[width=\linewidth]{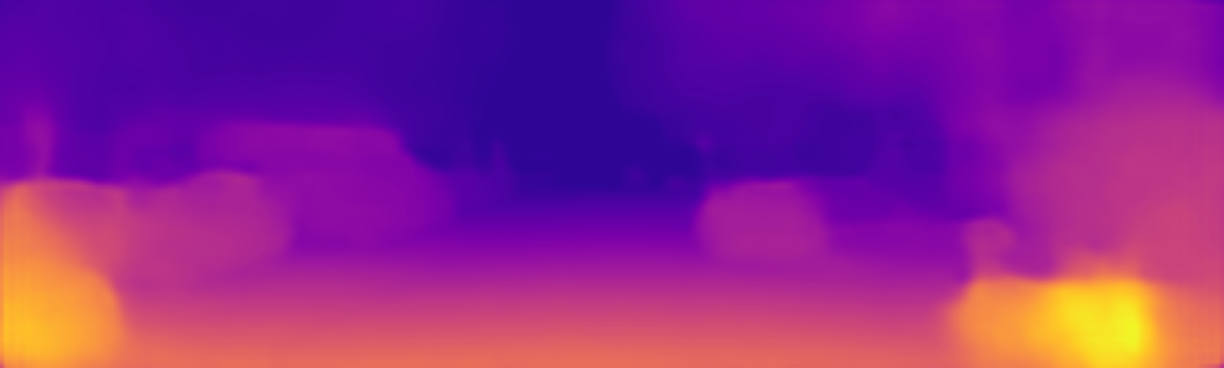}
  \end{subfigure}
    \begin{subfigure}[b]{0.12\linewidth}
    \includegraphics[width=\linewidth]{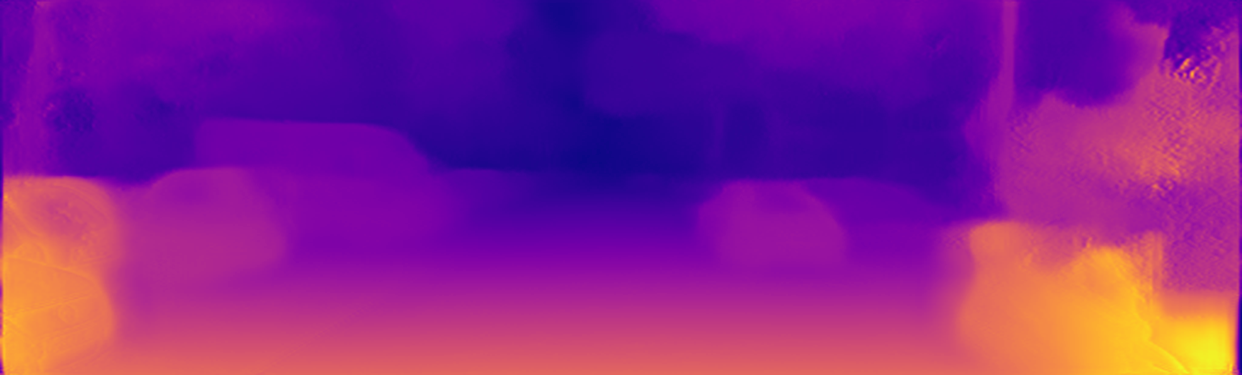}
  \end{subfigure}
    \begin{subfigure}[b]{0.12\linewidth}
    \includegraphics[width=\linewidth]{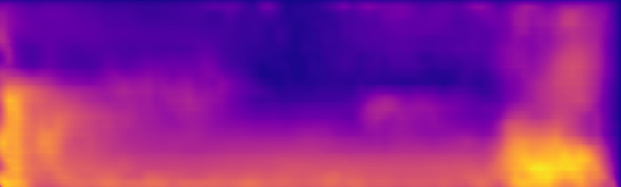}
  \end{subfigure}
      \begin{subfigure}[b]{0.12\linewidth}
    \includegraphics[width=\linewidth]{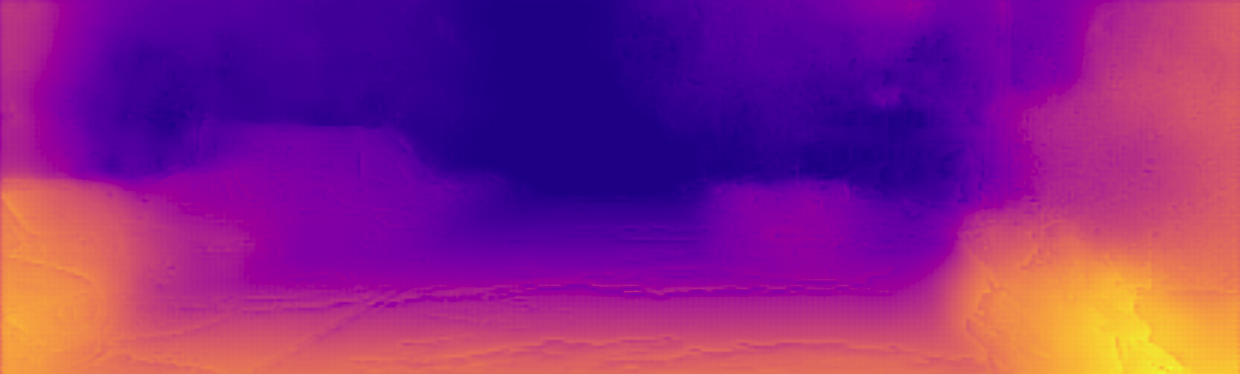}
  \end{subfigure}
    \begin{subfigure}[b]{0.12\linewidth}
    \includegraphics[width=\linewidth]{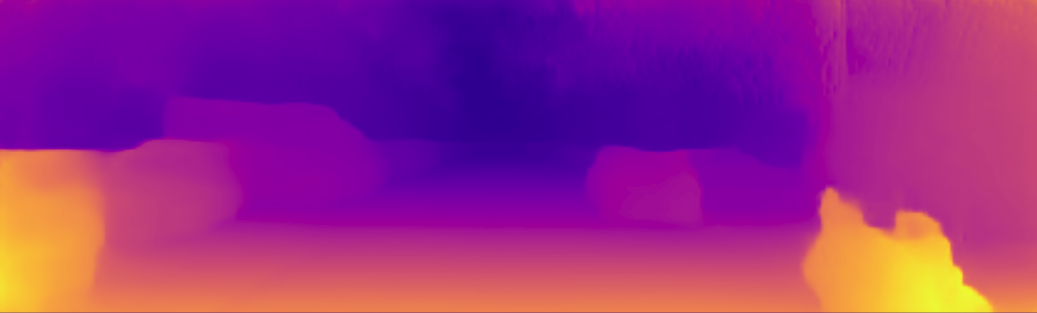}
  \end{subfigure}

    \begin{subfigure}[b]{0.12\linewidth}
    \includegraphics[width=\linewidth]{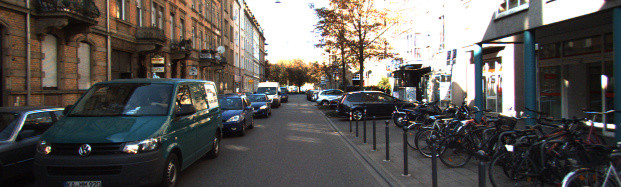}
  \end{subfigure}
  \begin{subfigure}[b]{0.12\linewidth}
    \includegraphics[width=\linewidth]{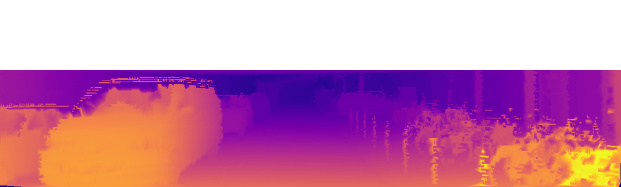}
  \end{subfigure}
    \begin{subfigure}[b]{0.12\linewidth}
    \includegraphics[width=\linewidth]{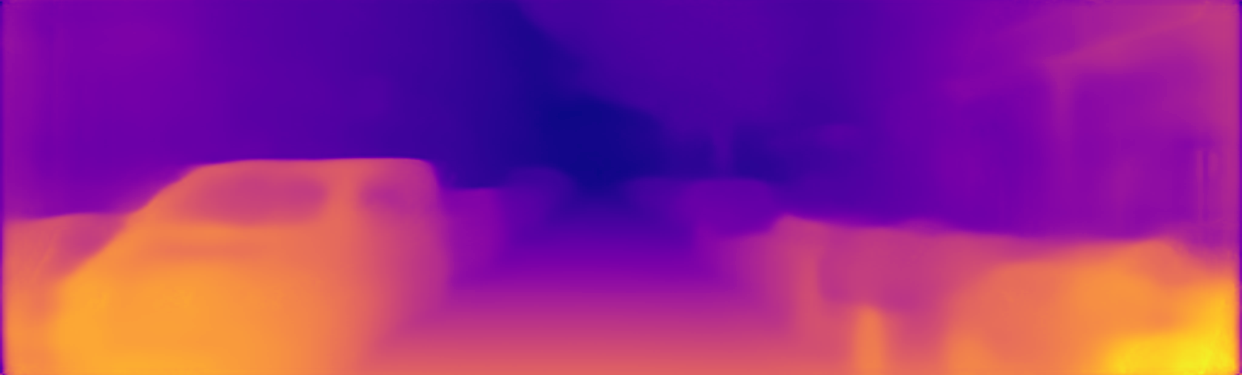}
  \end{subfigure}
    \begin{subfigure}[b]{0.12\linewidth}
    \includegraphics[width=\linewidth]{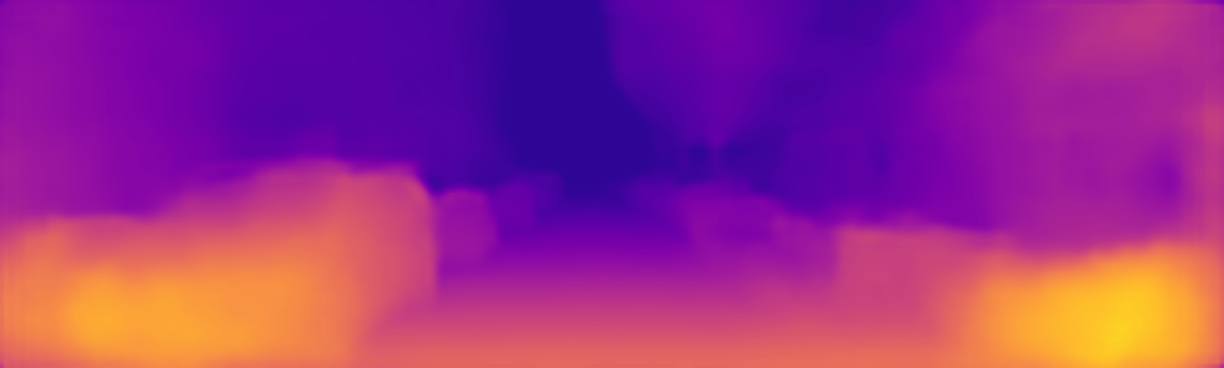}
  \end{subfigure}
    \begin{subfigure}[b]{0.12\linewidth}
    \includegraphics[width=\linewidth]{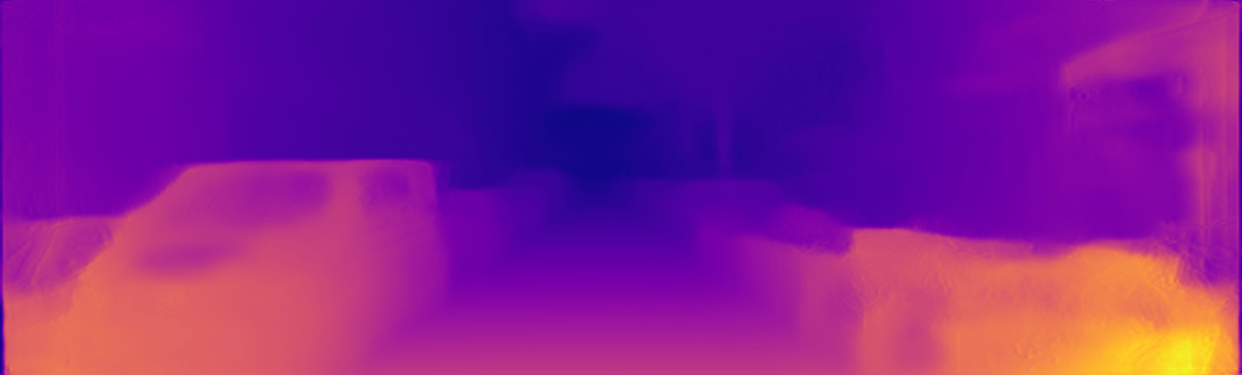}
  \end{subfigure}
    \begin{subfigure}[b]{0.12\linewidth}
    \includegraphics[width=\linewidth]{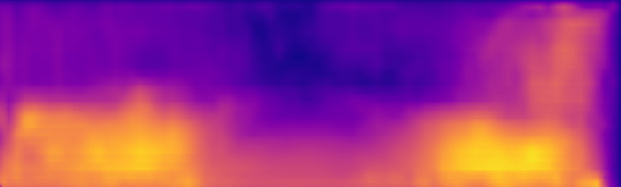}
  \end{subfigure}
      \begin{subfigure}[b]{0.12\linewidth}
    \includegraphics[width=\linewidth]{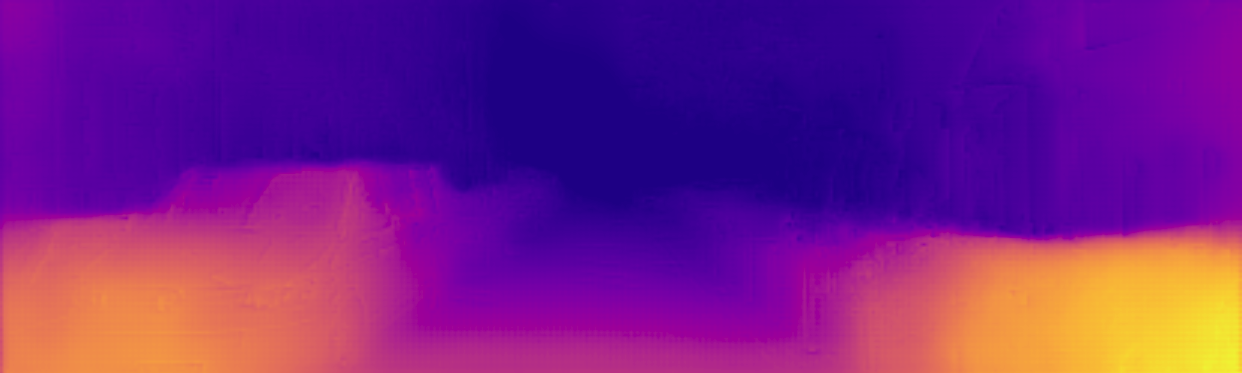}
  \end{subfigure}
    \begin{subfigure}[b]{0.12\linewidth}
    \includegraphics[width=\linewidth]{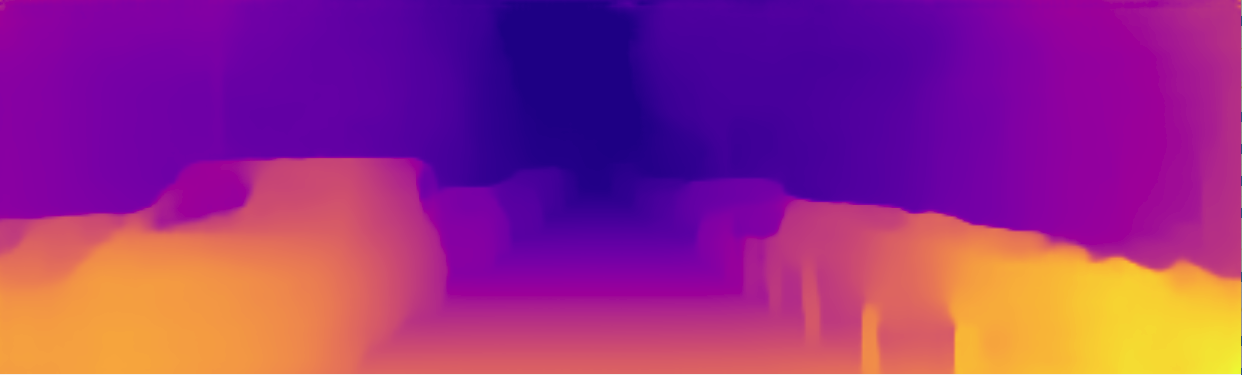}
  \end{subfigure}  
  \caption{Visual comparison between the state-of-the-art methods. For visualization the groundtruth depth is interpolated. Our method captures more details in thin structures, such as the motorcycle and columns in the lower right corner of figure rows 2 and 3.}
  \label{fig:depth_comp}
\end{figure*}

\begin{table*}[ht!]
\centering
\small
\resizebox{\linewidth}{!}{
\begin{tabular}{c|c|c c|c|c|c|c|c|c|c}%{ |p{3cm}||p{1.5cm}p{1.5cm}||p{1.2cm}||p{1.2cm}|p{1.1cm}|p{1.0cm}|p{1.6cm}| p{1.1cm}|p{1.0cm}|p{1.6cm}| }
 \thickhline
 Methods & Dataset & \multicolumn{2}{c|}{Supervised}  & \multicolumn{4}{c|}{Error metric} &\multicolumn{3}{c}{Accuracy metric} \\
\cline{3-11}
 & & depth&pose & RMSE& RMSE log  & Abs Rel & Sq Rel  & $\delta<1.25 $&$\delta<1.25^2$ & $\delta<1.25^3$\\
 \hline
 Zhou \textit{et al.} \cite{zhou2017unsupervised} &CS+K & & & 6.709 & 0.270 &0.183 & 1.595  & 0.734&0.902 &0.959\\
 Liu \textit{et al}. \cite{liu2015deep} & K & \checkmark & & 6.523 & 0.275 &   0.202  &  1.614   & 0.678 &0.895&0.965\\
  Eigen \textit{et al}. \cite{eigen2015predicting} & K &\checkmark &  & 6.307 & 0.282 & 0.203  & 1.548   & 0.702 & 0.890&0.958\\
  Yin \textit{et al.}\cite{Yin_2018_CVPR}& K& &  & 5.857 &0.233 & 0.155 & 1.296  &0.806 & 0.931&0.931 \\
 Zhan \textit{et al.} \cite{zhan2018unsupervised} &K & &\checkmark & 5.585 &  0.229 &0.135 & 1.132  & 0.820 & 0.933 & 0.971\\
Zou \textit{et al.} \cite{Zou_2018_ECCV} & K& &  & 5.507 & 0.223 & 0.150 & 1.124  &0.793 &  0.933&0.973 \\
 Godard \textit{et al.} \cite{godard2017unsupervised} &CS+K & &\checkmark &  5.311 & 0.219 & 0.124 & 1.076  &0.847
 &0.942 &0.973\\
  Atapour \textit{et al.} \cite{atapour2018real}& K+S* &\checkmark &  & 4.726 &  0.194 & 0.110 &  0.929  &  0.923 &0.967 & 0.984\\
 Kuznietsov \textit{et al.} \cite{kuznietsov2017semi}& K&\checkmark & \checkmark & 4.621 & 0.189 & 0.113 & 0.741  &0.875
&0.964 &0.988\\
  Yang \textit{et al.} \cite{yang2018deep}& K&\checkmark &  & 4.442 & 0.187 & 0.097 & 0.734  & 0.888 & 0.958 & 0.980 \\
  Fu \textit{et al.} (ResNet)  \cite{fu2018deep}& K&\checkmark &  & 2.727 &\underline{0.120} & \textbf{0.072} & 0.307  & \underline{0.932} &\underline{0.984}&0.994 \\
 \hline
 \hline
 \textbf{Ours-unsup} (multi-view) &K & &  & 2.320 & 0.153 &  0.112 & 0.418 & 0.882&0.974 &0.992\\ 
 \textbf{Ours-sup} (single-view)  &K&\checkmark &  & \underline{1.949} & 0.127&  0.088& \underline{0.245} & 0.915 &\underline{0.984} &\underline{0.996} \\
 \textbf{Ours-sup} (multi-view)&K&\checkmark &  & \textbf{1.698} & \textbf{0.110}&  \underline{0.077}& \textbf{0.205} & \textbf{0.941} &\textbf{0.990} &\textbf{0.998}\\
 \hline
\end{tabular}
}
\caption{Quantitative comparison of our network with other state-of-the-art CNN-based methods on KITTI \cite{Geiger2012CVPR} dataset using the Eigen Split \cite{eigen2015predicting}. \textit{Ours sup (Single-view)} is the evaluation of single view depth estimation result. \textit{Ours sup (mult-view)} is the result generated with the assistance of nine previous views. Even though our method is not restricted to a fixed number of frames per sequence during prediction or evaluation, we still use 10-frame sequence here for the consistency with the training. We discuss continuous estimation results in the ablation study Section \ref{sec:abl}. The bold numbers are results that rank first and the underlined results those that rank second. All results are capped at 80m depth.}
\label{table:depth}
\end{table*}

\section{Experiments}

In this section we show a series of experiments using the KITTI driving dataset \cite{Geiger2013IJRR,Geiger2012CVPR} to evaluate the performance of our RNN-based depth and visual odometry estimation method. As mentioned in Section \ref{sec:intro}, our architecture can be trained in a supervised or unsupervised mode. Therefore, we evaluated both supervised and unsupervised versions of our framework. In the following experiments we named the supervised version as \textit{ours-sup} and the unsupervised version as \textit{ours-unsup}. 
We also performed detailed ablation studies to show the impact of the different constraints, architecture choices, and estimations at different time-steps. 

%Our method evaluations are on multiple datasets and by comparisons against the state-of-the-art learning-based depth-estimation methods. 

\subsection{Implementation Details}
We set the weights for depth loss, smoothness loss, forward-backward consistency loss, and mask regularization to 1.0, 1.0, 0.05, and 0.05, respectively. The weight for the image reprojection loss is $\frac{1}{2^\delta-1}$, where $\delta$ is the number of frame intervals between source and target frame. We use the Adam \cite{kingma2014adam} optimizer with $\beta_1=0.9$, $\beta_2=0.999$. The initial learning rate is 0.0002. The training process is very time-consuming for our multi-view depth and odometry estimation network. One strategy we use to speed up our training process, without losing accuracy, is first to pretrain the network with  the consecutive view reprojection loss for 20 epochs. Then we fine-tune the network with the multi-view reprojection loss for another 10 epochs.

% The detailed network architecture is shown in table \ref{table:netparam}

% \begin{table}[h!]
% \centering
% \resizebox{\linewidth}{!}{
% \begin{tabular}{ p{3cm}| p{2.5cm}|p{2.5cm} }
%  \hline
%  Type & Filters & Output size \\
%  \hline
%  Input &  &  128\times416\times3\\
% Conv\times2 &32@3\times3\times3 & 64\times208\times32 \\
% Conv\times2 &64@3\times3\times32 & 32\times104\times64\\ 
% Conv\times2 &128@3\times3\times64  & 16\times52\times128\\
% Conv\times2 &256@3\times3\times128 & 8\times26\times256\\
% Conv\times2 &256@3\times3\times256  & 4\times13\times256\\
% Conv\times2 &256@3\times3\times256  & 2\times7\times256\\
% Conv\times2 &512@3\times3\times256  & 1\times4\times512\\
% Deconv+ConvLSTM &256@3\times3\times512 & 2\times7\times256\\ 
% Deconv+ConvLSTM &128@3\times3\times256  & 4\times13\times128\\
% Deconv+ConvLSTM &128@3\times3\times128 & 8\times26\times128\\
% Deconv+ConvLSTM &128@3\times3\times128  & 16\times52\times128\\
% Deconv+ConvLSTM &64@3\times3\times128  & 32\times104\times64\\
% Deconv+ConvLSTM &32@3\times3\times64  & 64\times208\times32\\
% Deconv &16@3\times3\times32  & 128\times416\times16\\
% Conv (output) &1@3\times3\times16  & 128\times416\times1\\
% \hline
% \end{tabular}
% }
% \caption{Detailed network architecture. Conv$\times$2 represents two convolutions, the first one with stride 2 and second one with stride 1. All convolutions and decovolutions are followed by batch normalization and RELU activation. }
% \label{table:netparam}
% \end{table}

\subsection{Training datasets}

% \textbf{Indoor}. We used two publicly available datasets for indoor scenes. The first one is \textbf{SUN3D} \cite{xiao2013sun3d}, which is a large dataset with ground truth depth maps and camera poses. We selected 192 scenes from a total of 354 scenes as our training data. Then we randomly selected 30 scenes from the remaining 162 scenes as for validation and testing. The second dataset is \textbf{RGBD-SLAM} \cite{sturm12iros}, which is a smaller dataset but with higher camera pose accuracy. RGBD-SLAM provides a training and validation split of their dataset, so we directly used their split. 

We used the KITTI driving dataset \cite{Geiger2012CVPR} to evaluate our proposed framework. To perform a consistent comparison with existing methods, we used the Eigen Split approach \cite{eigen2015predicting} to train and evaluate our depth estimation network. From the 33 training scenes, we generated 45200 10-frame sequences. Here we used the stereo camera as two monocular cameras. A sequence of 10 frames contains either 10 left-camera or 10 right-camera frames. We resized the images from 375 $\times$ 1242  to 128$\times$416 for computational efficiency and to be comparable with existing methods.   The image reprojection loss is driven by motion parallax, so we discarded all static frames with baseline motion less than $\sigma=0.3$ meters during data preparation. 697 frames from the 28 test scenes were used for quantitative evaluation. For odometry evaluation we used the KITTI Odometry Split \cite{Geiger2012CVPR}, which contains 11 sequences with ground truth camera poses. We follow \cite{zhou2017unsupervised,Zhan_2018_CVPR}, which use sequences 00-08 for training and 09-10 as evaluation.

%In order to demonstrate the full capacity of our framework, the 9 previous frames of the 697 testing frames were used to form the 10-frame sequences and help generating depth for the testing frames. 

%  The \textbf{Make3D} \cite{saxena2009make3d} dataset was used to evaluate generalization. 

% We trained two separate models for the indoor and outdoor datasets. 

\subsection{Depth Estimation}\label{sec:compare}

To evaluate the depth estimation component of our multi-view depth and odometry network, we compare to the state-of-the-art CNN-based depth estimation methods.  Our network takes advantage of previous images and depths through recurrent units and thus achieves best performance when running on a continuous video sequence. However, it would be unfair to compare against single view methods when our method uses multiple views. On the other hand, if we also use only a single view for our method, then we fail to reveal the full capacity of our framework. Therefore, in order to present a more comprehensive depth evaluation, we report both our depth estimation results with and without previous views' assistance. \textit{Ours-sup (single-view)} is the single view (or first view) depth estimation result of our framework, which also shows the bootstrapping performance of our approach. \textit{Ours-sup (multi-view)} is the tenth view depth estimation result from our network. As shown in Table \ref{table:depth}, \textit{ours-sup (multi-view)} performs significantly better than all of the other supervised \cite{liu2015deep,eigen2015predicting,atapour2018real,yang2018deep,fu2018deep,kuznietsov2017semi} and unsupervised \cite{zhou2017unsupervised,Yin_2018_CVPR,Zou_2018_ECCV,Zhan_2018_CVPR,godard2017unsupervised} methods. The unsupervised version of our network outperforms the state-of-the-art unsupervised methods as well as several supervised methods. Both the supervised and unsupervised version of our network outperform the respective state-of-the-art by a large margin. Figure \ref{fig:depth_comp} shows a visual comparison of our method with other methods. Our method consistently captures more detailed structures, e.g., the motorcycle and columns in the lower right corner of the
figures in rows 2 and 3.

% \textbf{Our DenseSLAMNet} & K & \textbf{0.081} & \textbf{0.017}&  \textbf{2.924}& \textbf{0.112}

% \begin{figure}[h]
%     \centering
%     \includegraphics[width=12.5cm]{compare2.png}
%     \caption{Visual comparison between the results of Eigen \textit{et al}.\cite{eigen2015predicting} and ours on KITTI dataset \cite{Geiger2013IJRR}. Groundtruth depth is interpolated for visualization purpose.}
%     \label{KITTI}
% \end{figure}

\subsection{Pose Estimation}

We used the KITTI Odometry Split to evaluate our visual odometry network. For pose estimation we directly ran our method through the whole sequence instead of dividing into 10-frame sub-sequences. We compared to the state-of-the-art learning-based visual odometry methods \cite{Zhan_2018_CVPR,zhou2017unsupervised,Yin_2018_CVPR} as well as a popular monocular SLAM method: ORB-SLAM \cite{mur2015orb}. We used the KITTI Odometry evaluation criterion \cite{Geiger2012CVPR}, which computes the average translation and rotation errors over sub-sequences of length (100m, 200m, ... , 800m). %and the 5-frame absolute trajectory error criterion used by \cite{zhou2017unsupervised,Yin_2018_CVPR} for a comprehensive local and global trajectory evaluation. 

 \begin{table}[ht!]
\centering
\small
\resizebox{\linewidth}{!}{
\begin{tabular}{ |c||c c|c c| }
 \hline
 Methods & \multicolumn{2}{c|}{Seq 09} & \multicolumn{2}{c|}{Seq 10}\\
 &$t_{err}$(\%) & $r_{err}$(deg/m)&  $t_{err}$(\%) & $r_{err}$(deg/m)\\
  \hline
 ORB-SLAM \cite{mur2015orb} &15.30 &  0.003&  3.68 &  0.005\\
 GeoNet \cite{Yin_2018_CVPR} &43.76 & 0.160 &  35.60 & 0.138\\ Zhou \textit{et al.} \cite{zhou2017unsupervised} &17.84 & 0.068 &  37.91 & 0.178\\
  Zhan \textit{et al.}  \cite{Zhan_2018_CVPR} &11.92 & 0.036 &  12.62 & 0.034\\ 
   DeepVO \textit{et al.} \cite{wang2017deepvo} &- & - &  8.11 & 0.088\\ 
  \textbf{Our unsupervised} & 9.88 & 0.034 & 12.24 & 0.052\\
 \textbf{Our supervised} & 9.30 & 0.035 & 7.21 & 0.039\\ 
 \hline
\end{tabular}
}
\caption{Quantitative comparison of visual odometry results on the KITTI Odometry dataset. $t_{err}$ is the percentage of average translational error and $r_{err}$ is the average degree per meter rotational error.}
\label{table:traj}
\end{table}
  
  \begin{figure}[ht!]
\captionsetup[subfigure]{labelformat=empty}
  \centering
  \begin{subfigure}[b]{0.48\linewidth}
  \includegraphics[width=\linewidth]{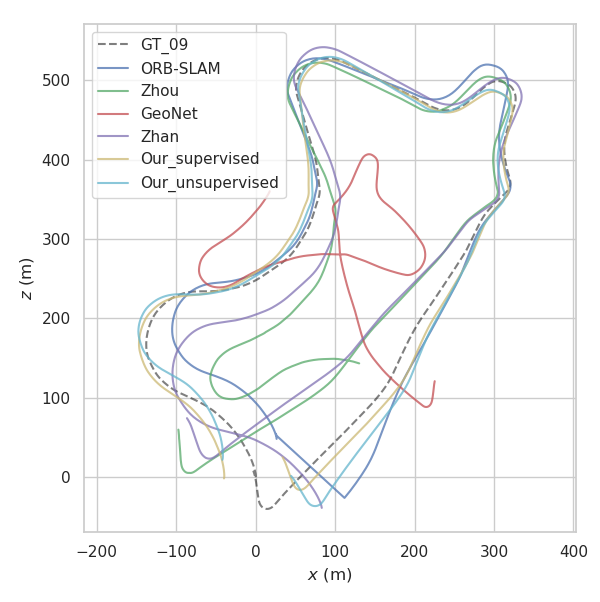}
  \end{subfigure}
  \begin{subfigure}[b]{0.48\linewidth}
  \includegraphics[width=\linewidth]{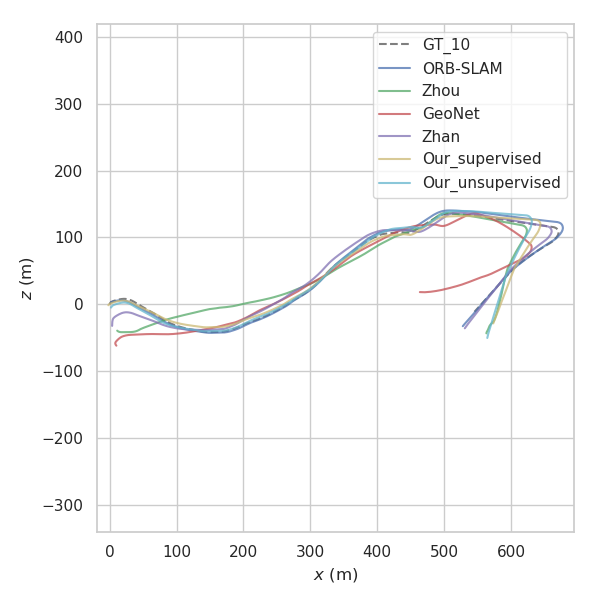}
  \end{subfigure}
  \caption{Visual comparison of full trajectories on Seq 09 (left) and 10 (right). Our predictions are closest to groundtruth (GT\_09 and GT\_10).}
  \label{fig:traj}
  \end{figure}

Both the monocular ORB-SLAM and the unsupervised learning-based visual odometry methods are suffering from scale ambiguity, so we aligned their trajectories with groundtruth prior to evaluation using evo\footnote{github.com/MichaelGrupp/evo}.  The supervised version of our method (absolute depth supervision) and the stereo supervised method \cite{Zhan_2018_CVPR} are able to estimate camera translations at absolute scale, so there is no post-processing for these two methods.

Table \ref{table:traj} shows quantitative comparison results based on the KITTI Visual Odometry criterion.  Figure \ref{fig:traj} shows a visual comparison of the full trajectories for all the methods. Including our method, all the full trajectories of learning-based visual odometry methods are produced by integrating frame-to-frame relative camera poses over the whole sequence without any drift correction.  

The methods \cite{zhou2017unsupervised,Yin_2018_CVPR} take a small sub-sequence  (5 frames) as input and estimate relative poses between frames within the sub-squence. There is no temporal correlation between different sub-sequences and thus the scales are different between those sub-sequences. However, our method can perform continuous camera pose estimation within a whole video sequence for arbitrary length. The temporal information is transmitted through recurrent units for arbitrary length and thus maintains a consistent scale within each full sequence.

%   \begin{figure}[ht!]
% \captionsetup[subfigure]{labelformat=empty}
%   \centering
%   \begin{subfigure}[b]{0.48\linewidth}
%   \caption{\scriptsize Seq 09}
%   \includegraphics[width=\linewidth]{images/09_trajectories_noscale}
%   \end{subfigure}
%   \begin{subfigure}[b]{0.48\linewidth}
%   \caption{\scriptsize Seq 10}
%   \includegraphics[width=\linewidth]{images/10_trajectories_noscale}
%   \end{subfigure}
%   \caption{Visual comparison of full trajectories on seq 09 and 10 without scale correction.}
%   \label{fig:traj_scale}
%   \end{figure}

% Figure \ref{fig:traj_scale} shows the full trajectories before scale correction. It can be seen that the trajectory shape of our method is very similar to the groundtruth trajectory which indicates that our prediction is only ambiguous to a single scale.

\subsection{Ablation study}
\label{sec:abl}

In this section we investigate the important components: placements of the recurrent units, multi-view reprojection and forward-backward consistency constraints in the proposed depth and visual odometry estimation network. 

\begin{figure}[t!]
%\captionsetup[subfigure]{labelformat=empty}
  \centering
  \begin{subfigure}[b]{0.32\linewidth}
  \includegraphics[width=\linewidth,height=1.5cm]{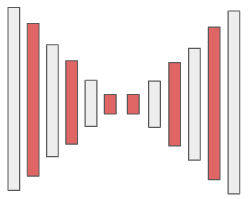}
  \caption{\scriptsize Full LSTM}
  \end{subfigure}
  \begin{subfigure}[b]{0.32\linewidth}
  \includegraphics[width=\linewidth,height=1.5cm]{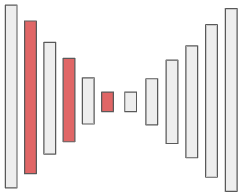}
  \caption{\scriptsize Encoder LSTM}
  \end{subfigure}
    \begin{subfigure}[b]{0.32\linewidth}
  \includegraphics[width=\linewidth,height=1.5cm]{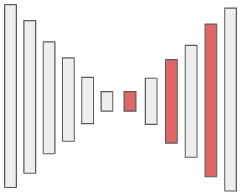}
  \caption{\scriptsize Decoder LSTM}
  \end{subfigure}
  \caption{Three different architectures depend on the placements of recurrent units. (a) We put a convolutional LSTM after every convolution or deconvolution layer. (b) We only place convolutional LSTM in the encoder. (c) We only place convolutional LSTM in the decoder.}
  \label{fig:arch_ablation}
  \end{figure}

\textbf{Placements of recurrent units.} Convolutional LSTM units are essential components for our framework to leverage temporal information in depth and visual odometry estimation. Thus we performed a series of experiments to demonstrate the influence of these recurrent units as well as the choice for the placements of recurrent units in the network architecture. We tested three different architecture choices which are shown in Figure \ref{fig:arch_ablation}. The first one is interleaving LSTM units across the whole network (full LSTM). The second one is interleaving LSTM units across the encoder (encoder LSTM). The third one is interleaving LSTM units across the decoder (decoder LSTM). Table \ref{table:arc_comp} shows the quantitative comparison results. It can be seen that the encoder LSTM performs significantly better than the full LSTM and the decoder LSTM. Therefore, we chose the encoder LSTM as our depth estimation network architecture.
  
\begin{table}[t!]
\centering
\small
\begin{tabular}{|l|c|c|c|c|}
\hline
Method & RMSE& RMSE log &Abs Rel & Sq Rel \\
\hline\hline
full LSTM &  1.764 & 0.112 &  0.079 & 0.214\\
decoder LSTM & 1.808 &  0.117 &  0.082 & 0.226\\
encoder LSTM & \textbf{1.698} & \textbf{0.110} &  \textbf{0.077} &\textbf{0.205}\\

\hline
\end{tabular}
\caption{Ablation study on network architectures. The evaluation data and protocol are the same as table \ref{table:depth}. }
\label{table:arc_comp}
\end{table}

\begin{figure}[ht!]
%\captionsetup[subfigure]{labelformat=empty}
  \centering
  \begin{subfigure}[b]{0.32\linewidth}
  \includegraphics[width=\linewidth]{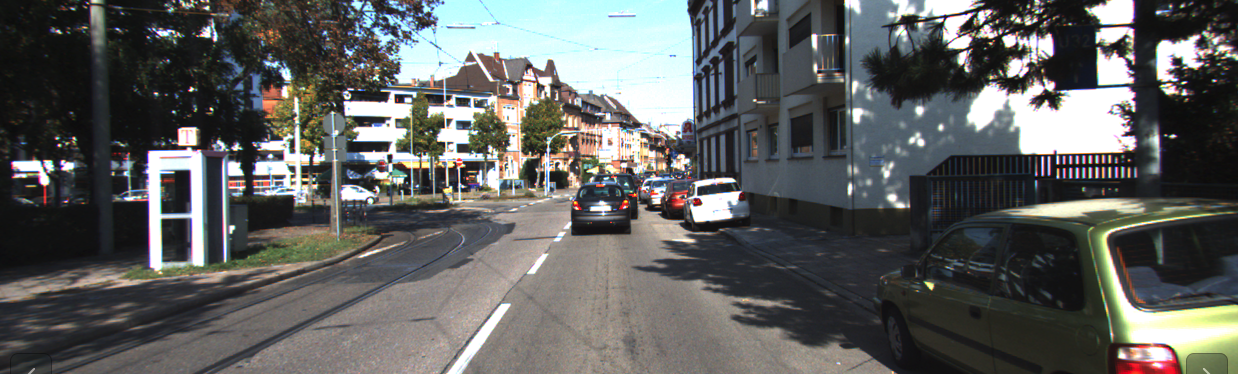}
  \end{subfigure}
  \begin{subfigure}[b]{0.32\linewidth}
  \includegraphics[width=\linewidth]{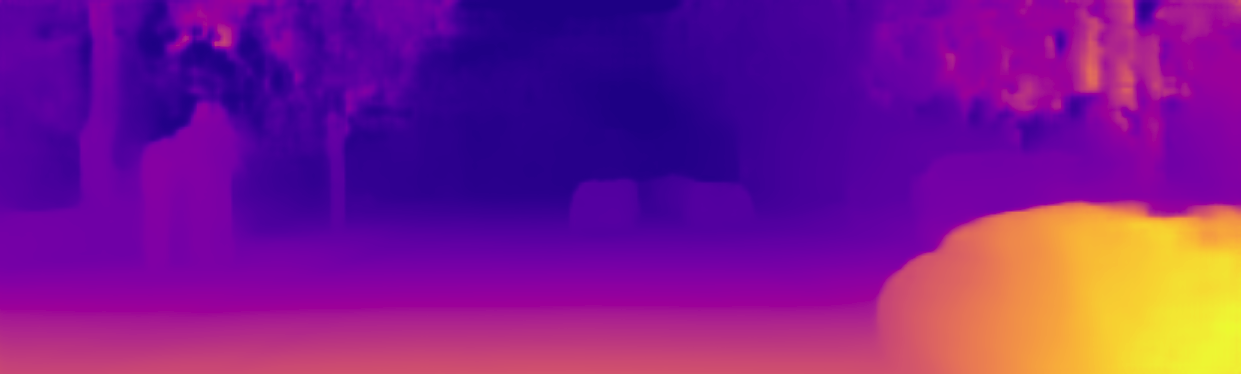}
  \end{subfigure}
    \begin{subfigure}[b]{0.32\linewidth}
  \includegraphics[width=\linewidth]{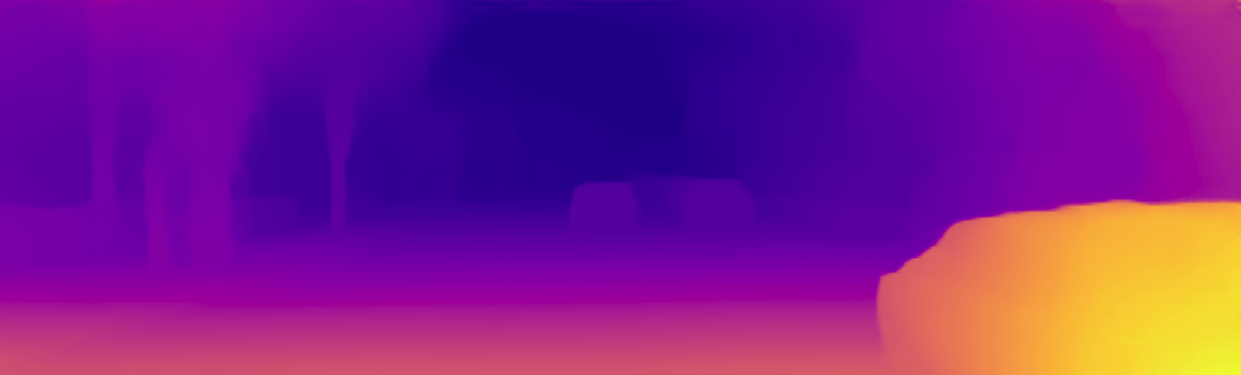}
  \end{subfigure}
  
  \begin{subfigure}[b]{0.32\linewidth}
  \includegraphics[width=\linewidth]{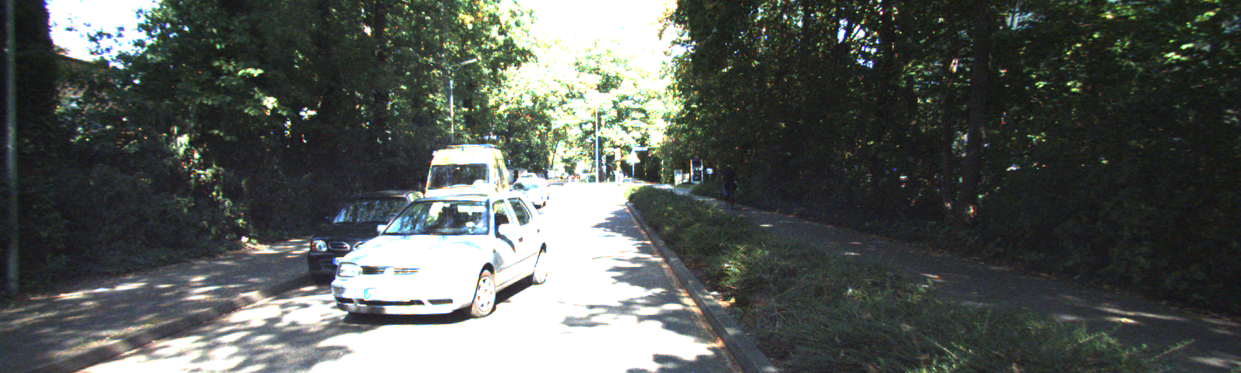}
  \caption{\scriptsize Input}
  \end{subfigure}
  \begin{subfigure}[b]{0.32\linewidth}
  \includegraphics[width=\linewidth]{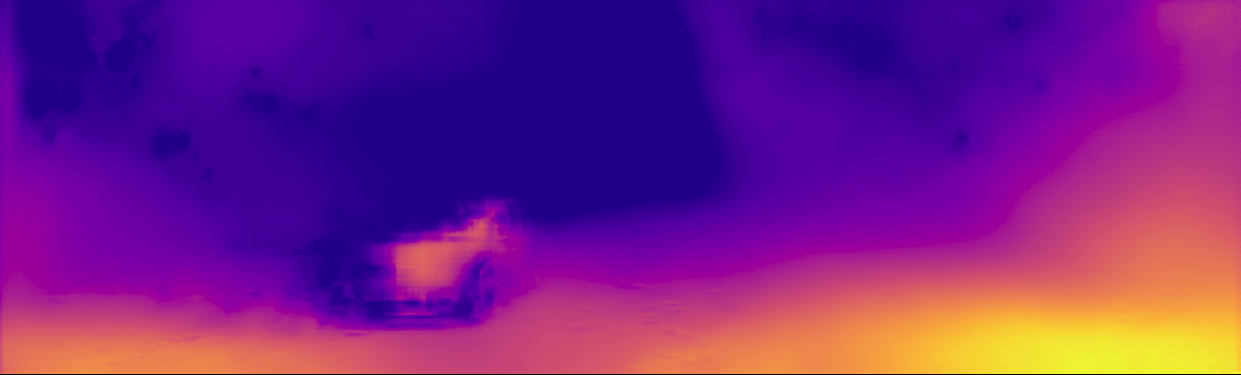}
  \caption{\scriptsize Consecutive reproj.}
  \end{subfigure}
  \begin{subfigure}[b]{0.32\linewidth}
  \includegraphics[width=\linewidth]{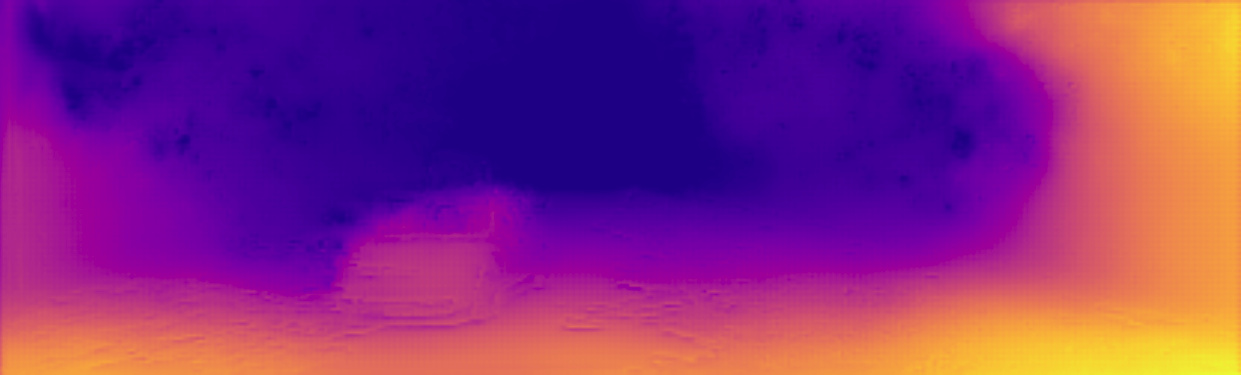}
  \caption{\scriptsize Muti-view reproj.}
  \end{subfigure}
  
  \caption{Visual examples between networks trained using consecutive image reprojection loss and those using multi-view reprojection loss. Results in the first row are from \textit{ours-sup}, and results in the second row are from \textit{ours-unsup}.}
  \label{fig:loss}
  \end{figure}

\textbf{Multi-view reprojection and forward-backward consistency constraints.}  To investigate the performance gain from the multi-view reprojection and forward-backward consistency constraints, we conducted another group of experiments. Table \ref{table:loss} shows the quantitative evaluation results. We compared among three methods: with only the consecutive image reprojection constraint (\textit{Ours-d}), with the consecutive image reprojection constraint and the forward-backward consistency constraint (\textit{Ours-dc}), and with the multi-view reprojection constraint and the forward-backward consistency constraint (\textit{Ours-mc}). The multi-view reprojection loss is more important in the unsupervised training, which is shown by the results of the last two rows in Table \ref{table:loss}. Figure \ref{fig:loss} shows a qualitative comparison between networks trained using consecutive image reprojection loss and those using multi-view reprojection loss. It can be seen that multi-view reprojection loss provides better supervision to areas that lack groundtruth depth.

\begin{table}[h!]
\centering
\small
\begin{tabular}{|l|c|c|c|c|}
\hline
Method & RMSE& RMSE log &Abs Rel & Sq Rel \\
\hline\hline
Ours-d &  1.785 & 0.116 &  0.081 & 0.214\\
Ours-dc & 1.759 & 0.113 &  0.079 & 0.215\\
Ours-mc & 1.698 & 0.110&  0.077& 0.205\\

\hline\hline
Ours-dc unsup &  2.689 & 0.184 &  0.138 &0.474\\
Ours-mc unsup& 2.361 & 0.157 &  0.112 &0.416\\
\hline
\end{tabular}

\caption{Ablation study on multi-view reprojection and forward-backward flow consistency constraints. \textit{d} stands for consecutive image reprojection.  \textit{m} stands for multi-view image reprojection. \textit{c} stands for forward-backward flow consistency constraint. The first three rows are comparison between supervised training and the last two rows are unsupervised. }
\label{table:loss}
\end{table}

\begin{table}[h!]
\centering
\small
\resizebox{\linewidth}{!}{
\begin{tabular}{|c|c|c|c|c|}
\hline
Window size & RMSE& RMSE log &Abs Rel & Sq Rel \\
\hline\hline
1 &  1.949 & 0.127 &  0.088 &0.245\\
3& 1.707 & 0.110 &  0.077 &0.206\\
5 & 1.699 &  0.110 &  0.077 &0.205\\
10 & 1.698 &  0.110 &  0.077 &0.205\\
20 & 1.711 &  0.117 &  0.077 &0.208\\
Whole seq. & 1.748 &  0.119 &  0.079 &0.214\\
\hline
\end{tabular}
}
\caption{Depth estimation with different time-window sizes. }
\label{table:time}
\end{table}

\textbf{Estimation with different temporal-window sizes.} Table \ref{table:time} shows a comparison between depth estimation with different temporal-window sizes, i.e., the number of frames forming the temporal summary. Here we use the Eigen Split 697 testing frames for these sliding-window-based evaluations. In addition, we also ran through each whole testing sequence and again performed evaluation on those 697 testing frames. The result demonstrates that 1) the performance of the depth estimation is increasing with the number of depth estimations performed before the current estimation; 2) the performance of the depth estimation is not increasing after 10 frames; 3) even though our network is trained on 10-frame based sub-sequences, it can succeed on an arbitrary length sequences.

%-------------------------------------------------------------------------
\subsection{Conclusion}

 In this paper we presented an RNN-based, multi-view method for depth and camera pose estimation from monocular video sequences. We demonstrated that our method can be trained either supervised or unsupervised and that both produce superior results compared to the state-of-the-art in  learning-based depth and visual odometry estimation methods. Our novel network architecture and the novel multi-vew reprojection and forward-backward consistency constraints let our system effectively utilize the temporal information from previous frames for current frame depth and camera pose estimation. In addition, we have shown that our method can run on an arbitrary length video sequences while producing temporally coherent results. 
 
%In the future we want to take the scene dynamic, occlusion, and illumination changes into account and explore the possibility to apply our framework on more difficult dataset such as endoscopic video sequences. For the visual odometry, even though we have taken multiple frames into account, there is no drift correction or loop closure integrated. Therefore, using a third network to explicitly optimize the full camera trajectory and perform loop closure will be an interesting direction and make the whole system more comparable with the state-of-the-art SLAM systems. 

{\small
\bibliographystyle{ieee}
\bibliography{egbib}
}

\end{document}